\documentclass[letterpaper, 10 pt, conference]{ieeeconf}  

\IEEEoverridecommandlockouts                              

\usepackage{graphicx} 
\usepackage{amsmath} 
\usepackage{listings}
\usepackage{algorithm}
\usepackage{algpseudocode}
\usepackage{textcomp}
\title{\LARGE \bf
A C Code Generator for Fast Inference and Simple Deployment of Convolutional Neural Networks on Resource Constrained Systems
}

\author{Oliver Urbann$^{1}$, Simon Camphausen, Arne Moos, Ingmar Schwarz, S\"oren Kerner, Maximilian Otten 
\thanks{The work of Oliver Urbann has been funded by the Federal Ministry of Education and Research of Germany as part of the competence center for machine learning ML2R (01-S18038A).}
\thanks{$^{1}$Oliver Urbann is with Automation and embedded Systems,
       Fraunhofer IML, 44227 Dortmund, Germany
        {\tt\small oliver.urbann@iml.fraunhofer.de}}%
}

\begin{document}

\maketitle
\thispagestyle{empty}
\pagestyle{empty}

\begin{abstract}
Inference of Convolutional Neural Networks in time critical applications usually requires a GPU. In robotics or embedded devices these are often not available due to energy, space and cost constraints. Furthermore, installation of a deep learning framework or even a native compiler on the target platform is not possible. This paper presents a neural network code generator (NNCG) that generates from a trained CNN a plain ANSI C code file that encapsulates the inference in single a function. It can easily be included in existing projects and due to lack of dependencies, cross compilation is usually possible. Additionally, the code generation is optimized based on the known trained CNN and target platform following four design principles. The system is evaluated utilizing small CNN designed for this application. Compared to TensorFlow XLA and Glow speed-ups of up to 11.81 can be shown and even GPUs are outperformed regarding latency.
\end{abstract}

\section{INTRODUCTION}

\subsection{Motivation and Related Work}

Recent scientific advancements have led to a general acceptance of various classes of deep learning architectures as state of the art in machine learning, e.g. Convolutional Neural Networks (CNN). Research focused previously on image classification with a large number of classes~\cite{ILSVRC15} and is currently shifting towards object detection with approaches like R-CNN~\cite{girshick2014rich}, where from the image region proposals are extracted and afterwards computed and classified utilizing CNN and SVM respectively.

CNN are thus a core component of a wide area of computer vision algorithms and are computationally expensive, usually accelerated by GPUs or FPGAs.
However, recent advancements in mobile autonomous robotics as well as the Internet of Things (IoT) has opened a wide area of highly promising applications for these kind of algorithms, in which GPUs are not available and optimization of algorithmic performance becomes essential to save energy.
At the same time those practical applications often come with the consequence that a large number of classes is not required and thus small CNN architectures are sufficient.

The first goal of this work is therefore to speed up the inference of small pretrained networks.
In general the motivation of this is twofold. First, the usual meaning is a high throughput given a large set of images to be classified. Overhead due to e.g. initialization is negligible. Second, and this is more important in this scope of application, the reduction of execution time - which correlates with latency as well as energy consumption.
In mobile robotic applications latency is important for near-real-time reaction to sudden changes in the environment.
In this case the latency should be as low as possible where computational time also correlates with energy savings.
While the energy consumption is also a factor in mobile robotic applications its way more important in the IoT application, where it single handedly defines the lifetime of such a device.
The set of images that must be classified at one time is rather low and thus the throughput is an suboptimal criteria for performance in this case.

The second aspect is target platform deployability. Typically the network is embedded in a framework that provides images and processes the results. Embedding machine learning frameworks like TensorFlow~\cite{tensorflow2015-whitepaper} or Caffe~\cite{Jia:2014:CCA:2647868.2654889} requires much overhead for an inference of a pretrained network.
As a consequence, tools for generating object code for inference were developed for those kind of frameworks where TensorFlow XLA\footnote{https://www.tensorflow.org/xla/} and Glow\footnote{https://facebook.ai/developers/tools/glow} are currently state-of-the-art.
But their applicability to generic target platforms is limited.
TensorFlow XLA generates object code that depends on TensorFlow code limiting the cross compilation capabilities for target platforms, whereas Glow's capabilities are currently limited to x86-64 and ARM64. It does not offer switches for other platforms, e.g. 32 bit targets, out of the box. 

\subsection{Contribution}
\label{sec:contrib}
In this paper we propose a neural network code generator (NNCG) that generates C code from a trained CNN model.
It focuses on the two relevant goals motivated previously:

\begin{itemize}
	\item Generic scope of applicability and cross compilation for various target platforms
	\item Generation of fast executables allowing CNN inference on resource constrained systems (small robots, embedded microcontrollers etc.) on a CPU only
\end{itemize}

\subsection*{Generic Deployment}
In contrast to common approaches which compile library code (e.g. Eigen) for operations like matrix multiplication and generate object code, we propose to generate plain ANSI C code.
Since specialized code for each atomic operation (e.g. multiplication or addition) is generated and weights are included as constants, no libraries or prewritten code are needed except math.h and libmath.

If desired target architecture dependent enhancements (e.g. SIMD instructions like SSE) can be utilized as well.
As a result, the code can easily be compiled using a cross compiler or natively compiled on any target platform.

\subsection*{Fast Executables}
\label{sec:introfe}

Utilizing a math library and compiling to object code relies on a good optimization performance of the compiler as well as on the efficiency of the library.
However, both library and optimizer are developed for any generic mathematical case.
Instead, we exploit our knowledge about CNN in general and especially for the specific trained model to generate the most optimal code.
Additionally, we intentionally choose C as output to fully benefit from the optimization capabilities of the compiler by generating code that is easy to optimize.

We identified four design principles to achieve those ideas, which we will discuss in detail in the following Sec. \ref{sec:NNCG}:

\begin{itemize}
	\item Loop unrolling and caching
	\item Conditional moves instead of branching
	\item Constants wherever possible
	\item Identification of applicable data structures for SIMD instructions
\end{itemize}

Usually the compiler should be able to handle most of these topics by itself. However, as the compiler has no background information this frequently fails in the field. 
It has to be noted that these design principles limit the application of NNCG to small networks, as loop unrolling and floats written in ASCII text lead to large code files. E.g. MobileNetV2~\cite{sandler2018mobilenetv2} would require approx. 4 MB only for printing all weights in ASCII which leads to C files difficult to handle for a compiler.

In the evaluation in Sec. \ref{sec:evaluation} we will show the advantage of addressing these points in NNCG based on small CNN adequate for our purpose.
We compare the performance of NNGC with both above mentioned tools (TF XLA and Glow), wherever possible on a PC as well as a mobile robotic platform.
We are able to show speed-ups of factor 1.41 up to 11.81 depending on network size and platform.
We also compare the latency of a system with and without GPU. We can show that with small networks and a small number of images to classify the latency of our executable is many times smaller.

\subsection{Structure}

The following Sec. \ref{sec:NNCG} describes the conceptual details of the NNCG and its implementation.
Afterwards the results of NNCG are compared to the current state-of-the-art on various target platforms, in Sec. \ref{sec:evaluation}.
This evaluation will focus on mobile robotics as an application area, since in general it offers a larger variety of pattforms with different computational performance levels. 
The final Sec. \ref{sec:conclusion} concludes the paper and provides an outlook to future research.

\section{Neural Network Code Generator (NNCG)}
\label{sec:NNCG}

In this section we first describe the design principles (see Sec.~\ref{sec:introfe}) in detail and continue how the CNN layer are realized fulfilling these principles.

The basic concept is the generation of C code from a trained Keras\footnote{https://keras.io} model during an exemplary classification of an image. We reimplemented various CNN layer (all layer required for a custom YOLO~\cite{DBLP:conf/cvpr/RedmonF17} net)  with focus on simplicity.

During the calculation of each layer C code is written for all atomic operations, e.g. multiplication, addition, max operator etc. including the involved values as constants.

\subsection{Design Principles}
\label{sec:desprin}
\subsubsection{Loop unrolling and caching}
\label{sec:unroll}
In general, a loop consists of code for checking if a condition is met to continue executing the loop and a branch that repeats the loop. This has (mainly) two disadvantages: (1) Code for condition checking and branching and (2) negative effects on the pipeline of the processor resulting in a pipeline filled with wrong instructions if the  processor cannot predict the condition correctly.

To mitigate this a compiler can unroll loops meaning the body of the loop is executed multiple times and the condition check and branch is thus executed less often. However, for this to work efficiently the number of loop iterations must be known, or further code is required to met the exact number of iterations.

On the other hand, unrolling results in more instructions that must be loaded from RAM which also affects the efficiency of the CPU cache.
If all loops are unrolled completely, all instruction are only executed once and thus caching cannot increase execution speed.

Thus, we organize loop unrolling in different levels so that it can be chosen depending on the cache architecture of the target platform and the structure of the CNN. At level 0 all loops are unrolled. Level 1 does not unroll the outer most loop and so forth.

\subsubsection{Conditional Moves}

A typical operation is to copy a value into a register under some condition. In higher programming languages this usually is realized by a conditionally executed code block with a copy. It is skipped if the condition is not met again resulting in the clearance of the pipeline.

Thus, common processors implement copy instructions that are always executed but only actually copy the data if the condition is met. In worst case the time for executing this instruction is lost which is usually faster than refilling the pipeline.

Modern compiler should be able to identify candidates for a conditional move. However, as NNCG knows the semantics it can help by using the ternary operator known in C.

\subsubsection{Constants}

In common frameworks a CNN model is loaded into RAM during run-time and weights are passed to the calculation. The inference then must access these arrays using some addressing scheme. This may lead to unnecessary overhead as we can write the known constants into the corresponding line.

\subsubsection{SIMD Instructions}
\label{sec:simd}
Single Instruction Multiple Data (SIMD) instructions perform the same operation on multiple values and can thus speed-up the inference significantly. Modern compiler support these instructions but must be able to identify possible parallel calculations. To do so, the structure of the network must be known at compile time.

During code generation the structure of the calculations (matrix multiplications etc.) and the dimensions of vectors and matrices are known. Thus, parallel structures can be identified and SIMD instructions generated.

But, SIMD instructions are platform dependent. Currently we support Intel's SSSE3 and a general architecture without platform dependent code. However, other platform specific optimizations, such as for ARM's Cortex-M~\cite{lai2018cmsis}, can be integrated into the code generator as well.

\subsection{Layer}

We focus our work on layers required to implement a small YOLO~\cite{DBLP:conf/cvpr/RedmonF17} net. The following layers are also sufficient for other small networks that are suitable for embedded systems.

\subsubsection{Convolution}

Convolutional layers are the most computational demanding layers and thus a focus of this work. We support zero-padding and strides. Possible activations functions are the softmax function and (leaky) ReLU which we describe later.

To support padding we set all values to zero that are out of bounds by defining
\begin{eqnarray}
\hat{x}_{ijk}=
\begin{cases}
x_{ijk}, & \text{if}\ 1\leq i \leq h_{in}\ \wedge \ 1 \leq j \leq w_{in}\\
0, & \text{otherwise},
\end{cases}
\end{eqnarray}
where $x_{ijk}$ is the input of the convolution layer as a scalar at $\left(i,j\right)$ and channel $k$, $h_{in}$ the height and $w_{in}$ the width. Applying this definition our implementation of the convolution  can be written as
\begin{eqnarray}
y_{ijk}&=&y_{ijk}+\sum_{n=1}^{h_k}\sum_{m=1}^{w_k}\sum_{o=1}^{c_{in}}\omega_{nmok}\hat{x}_{i+n-p_t,j+m-p_w,k}\nonumber\\
i &\in& \lbrace 1, 1+h_s, \dots, h_{in}-h_k+p_h+1 \rbrace \nonumber\\
j &\in& \lbrace  1, 1+w_s, \dots, w_{in}-w_k+p_w+1 \rbrace \nonumber\\
k &\in& \lbrace 1, \dots, c_{out}\rbrace,
\label{eq:conv}
\end{eqnarray}
with $y_{ijk}$ as the output at $\left(i,j\right)$ in channel $k$, $h_k$, $w_k$ height and width of the kernel, $c_{in}$ and $c_{out}$ the number of input/output channels, $\omega_{nmok}$ the kernel weight at $\left(n,m\right)$ for output channel $k$ and input channel $o$, $p_h$, $p_w$ the height and width of the padding, $h_s$, $w_s$ the step height and width and $h_{in}$, $w_{in}$ the dimensions of the input.

We see in Eq.~\ref{eq:conv} the calculation of a single value requires three nested loops. Furthermore, to calculate all output values three additional nested loops are required.
The implementation of the first design principle is thus a trade-off between loop unrolling and code size.
For the reasons explained in Sec.~\ref{sec:unroll} unrolling all loops infinitely is only adequate for small networks and thus we follow a configurable approach.
Currently, we support unrolling all loops with possible exceptions for the first and second outer loop and no unrolling.

To further specialize our code for different channel and spatial dimensions, we created multiple code versions of the convolution with different tradeoffs between cache utilization and register preassure. For each layer we independently benchmark every code version and select the one with the best runtime performance.

Implementation of design principle 3 depends on unrolling. If no loop is unrolled we generate an array containing all weights as constants. If loops are unrolled, the constants can be written into the corresponding code line.

For design principle 4 we identified the output channels (loop over $k$ in Eq.~\ref{eq:conv}) as a proper dimension for SIMD instructions. As can be seen, this loop does not affect the three inner loops and is thus simple to apply.
For SSSE3 the number of channels (in Eq.~\ref{eq:conv} denoted by $c_{out}$) should be dividable by 4 such that the number of filters in convolutional layers should be a multiple of 4.

\subsubsection{Max-Pooling}

The max-pooling layer searches for the maximum of all values in a two-dimensional window,
\begin{eqnarray}
y_{ijk}&=&\max\left({x_{i\cdot h_s,j\cdot w_s,k}, \dots, x_{i\cdot h_s+h_k,j\cdot w_s+w_k,k}}\right)\nonumber\\
c_{out}&=&c_{in} \nonumber\\
i &\in& \lbrace 1, h_{in} - h_k+1\rbrace\nonumber\\
j &\in& \lbrace 1, w_{in} - w_k+1\rbrace\nonumber\\
k &\in& \lbrace 1, \dots, c_{out}\rbrace.
\end{eqnarray}

This two-dimensional window requires (in a basic form) two nested loops with additional three outer loops for each value of the output feature maps. Comparable to the convolution layer, we support no unrolling and full unrolling with possible exceptions for the outermost and second outermost loop. Furthermore, SIMD instructions are applied over channels if the number of filters in the previous convolution layer is a multiple of 4 (SSSE3).
\subsubsection{(Leaky) ReLU}

The ReLU layer consists of only three nested loops. We apply the same rules for unrolling as for max-pooling. The implementation of ReLU is simply,
\begin{eqnarray}
y_{ijk}=\max\left(x_{ijk}, 0\right).
\end{eqnarray}
A leaky ReLU layer can mathematically described as:
\begin{eqnarray}
y_{ijk}=
\begin{cases}
x_{ijk}, & \text{if}\ 0 < x_{ijk}\\
\alpha \cdot x_{ijk}, & \text{otherwise}
\end{cases},
\end{eqnarray}

where $\alpha$ is a factor realizing the "leaky" feature. The implementation for SSSE3 is also a max function with additional code for $\alpha$. For a general architecture we utilize the conditional operator of the C language to implement the second design principle supporting the compiler utilizing conditional moves.

\subsubsection{Batch Normalization}

Batch Normalization was introduced to improve the performance of CNNs, as well as to stabilize the training process. The layer consists of a learnable affine transformation of the input feature map,

\begin{eqnarray}
y_{ijk}=\frac{x_{ijk}-\mu}{\sigma}.
\end{eqnarray}

The calculation can be incorporated into a preceeding convolutional layer by modifying the weights and bias as shown below,

\begin{align*}
\mathbf{bn}\left( \mathbf{conv}\left( x \right) \right) &= \frac{\sum_i x_iw_i - \mu}{\sigma} \\
&= \frac{\sum_i x_iw_i}{\sigma} - \frac{\mu}{\sigma} \\
&= \sum_i x_i \left(\frac{w_i}{\sigma}\right) - \left(\frac{\mu}{\sigma}\right).
\end{align*}

\section{Evaluation}
\label{sec:evaluation}
\begin{table}[ht]
	\caption{Ball classifier CNN.}
	\label{tab:ball}
	\begin{center}
		\begin{tabular}{lcccc}
			Layer       & \# & Size & Stride & Padding\\
			\hline
			Input       & 1  & 16x16&        &        \\
			Conv        & 8  & 5x5  &  2x2   & same   \\
			ReLU        &    &      &        &        \\
			Max-Pool    &    & 2x2  &  2x2   &        \\
			Conv        & 12 & 3x3  &        & valid  \\
			ReLU        &    &      &        &        \\
			Conv        & 2  & 2x2  &        & valid  \\
			Soft-Max    &    &      &        &        \\
		\end{tabular}
	\end{center}
\end{table}
\begin{table}[ht]
	\caption{Pedestrian classifier CNN.}
	\label{tab:ped}
	\begin{center}
		\begin{tabular}{lcccc}
			Layer       & \# & Size & Factor & Padding\\
			\hline
			Input       & 1  & 18x36&        &        \\
			Conv        & 12 & 3x3  &        & same   \\
			ReLU        &    &      &        &        \\
			Max-Pool    &    & 2x2  &        &        \\
			Conv        & 32 & 3x3  &        & same   \\
			Leaky-ReLU  &    &      & 0.1    &        \\
			Max-Pool    &    & 2x2  &        &        \\
			Conv        & 64 & 3x3  &        & same   \\
			Leaky-ReLU  &    &      & 0.1    &        \\
			Max-Pool    &    & 2x2  &        &        \\
			Dropout     &    &      & 0.3    &        \\
			Conv        & 2  & 4x2  &        & valid  \\
			Soft-Max    &    &      &        &        \\
		\end{tabular}
	\end{center}
\end{table}
\begin{table}[ht]
	\caption{Robot detector CNN.}
	\label{tab:robot}
	\begin{center}
		\begin{tabular}{lcccc}
			Layer       & \# & Size & Factor & Padding\\
			\hline
			Input       & 3  & 80x60&        &        \\
			Conv        & 8  & 3x3  &        & same   \\
			Batch Norm. &    &      &        &        \\
			Leaky ReLU  &    &      & 0.1    &        \\
			Max-Pool    &    & 2x2  &        &        \\
			Conv        & 12 & 3x3  &        & same   \\
			Batch Norm. &    &      &        &        \\
			Leaky-ReLU  &    &      & 0.1    &        \\
    		Conv        & 8  & 3x3  &        & same   \\
    		Batch Norm. &    &      &        &        \\
    		Leaky ReLU  &    &      & 0.1    &        \\
    		Max-Pool    &    & 2x2  &        &        \\
    		Conv        & 16 & 3x3  &        & same   \\
    		Batch Norm. &    &      &        &        \\
    		Leaky ReLU  &    &      & 0.1    &        \\
   			Conv        & 20 & 3x3  &        & same   \\
    		Batch Norm. &    &      &        &        \\
    		Leaky ReLU  &    &      & 0.1    &        \\

		\end{tabular}
	\end{center}
\end{table}
This section evaluates the main goals of the this work as mentioned in Sec.~\ref{sec:contrib}: simple deployment and fast executables. We compare NNCG with both tools mentioned in the introduction that have comparable intentions: TensorFlow XLA and Glow in versions available in December 2018, 1.12 and c27b61c respectively.

Common robotic platforms are based on CPUs at different performance level.
As an example for cognitive mobile robotic applications the Robocup Standard Plattform League has been chosen.
Its robot Nao by SoftBank Robotics\footnote{https://www.softbankrobotics.com/emea/en/nao} is a typical example of a small and cheap mobile robot, which intentionally lack a GPU to save energy.
But if energy consumption, heat dissipation and cost are relatively neglectable, also a GPU can be integrated. We thus include various target platforms in our evaluation. A desktop processor Intel i7 8650U with Ubuntu 14.04, an energy efficient platform Intel Atom J1900 with Ubuntu 14.04, the Nao V5 by SoftBank Robotics (Intel Atom Z530) with a custom 32 bit Linux and the NVIDIA GPU GTX 1050 in a mobile system.

We evaluate both goals by presenting exemplary scenarios in simple robotic example applications: a ball detector for robot soccer, a pedestrian detector and a robot detector, all inferred on the mentioned target platforms. The CNN utilized for these purposes are described in Tab.~\ref{tab:ball}, Tab.~\ref{tab:ped} and Tab.~\ref{tab:robot}, respectively. Our evaluation is based on custom CNN designed to be small enough to lead to acceptable sizes of the C code file which is also desirable in terms of inference speed on mobile platforms. For example, a MobileNet V2 leads to an 78 MB C code file. We are still able to compile and run this file. However, we suggest smaller networks and thus we evaluate utilizing the networks presented.
The CNN structure is chosen such that decent classification results can be achieved and the networks are adequate for a simple application on embedded devices. 

To evaluate the first goal of this work, we give a subjective and comparative overview about simplicity and applicability of the tools to generate an executable of the mentioned pretrained networks (ball and pedestrian detector).
Afterwards, the second goal is evaluated by comparing the time required to infer a single image on CPU and GPU using NNCG, TensorFlow XLA and Glow. Besides this we also show how single features of NNCG can lower the latency.
We are also interested in how a GPU could perform if no overhead is present. We thus additionally evaluate the throughput of the GPU by applying a large set of images on the GPU and compare this speed per image with the tools on other platforms.

\subsection{Training}

For each scenario we train the CNN presented above utilizing realistic datasets. 

\subsubsection*{Ball CNN}

\begin{figure}[thpb]
	\centering
	\includegraphics[scale=1.5]{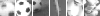}
	\caption{Three positive examples (left) and three negative examples (right) of the ball dataset.}
	\label{fig:ball_ex}
\end{figure}

The CNN presented in Tab.~\ref{tab:ball} is utilized in a pipeline for ball detection comparable to an R-CNN~\cite{girshick2014rich} in robot soccer. 
As a first step possible ball regions are extracted{~\cite{vision15}. For this the image is first traversed along scanlines and segmented. On the resulting ball segments, multiple scanlines are created to find ball edge points. These in turn are used for circle fitting leading to a ball candidate for the presented CNN which is used for feature extraction and verification. An average of 20 ball candidates is created per image.
  
The size of the CNN can be very small for multiple reasons. A ball is an object with high contrast (white with black spots) and the appearance is invariant with respect to orientation.

The dataset consist of 455107 images with 125615 balls at a resolution of 16x16, see Fig.~\ref{fig:ball_ex} for some examples. With 5\% of the images for evaluation our trained CNN has an accuracy of 99.975\%.

\subsubsection*{Pedestrian CNN}
\begin{figure}[thpb]
	\centering
	\includegraphics[scale=1.5]{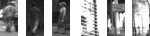}
	\caption{Three positive examples (left) and three negative examples (right) of the Daimler pedestrian dataset.}
	\label{fig:ped_ex}
\end{figure}

In real world scenarios pedestrian or human detection is an important application.
Humans are significantly harder to detect than a ball and we selected this as an example application to compare NNCG utilizing larger CNN, see Tab.~\ref{tab:ped}. For training we selected the Daimler pedestrian dataset~\cite{munder2006experimental}, which consist of 49000 images with 24000 images of humans at a resolution of 18x36 per image, see Fig.~\ref{fig:ped_ex} for example images.
With 10\% of the images for validation we achieve an accuracy of 99.02\%.

\subsubsection*{Robot detector}

\begin{figure}[thpb]
	\centering
	\includegraphics[scale=0.25]{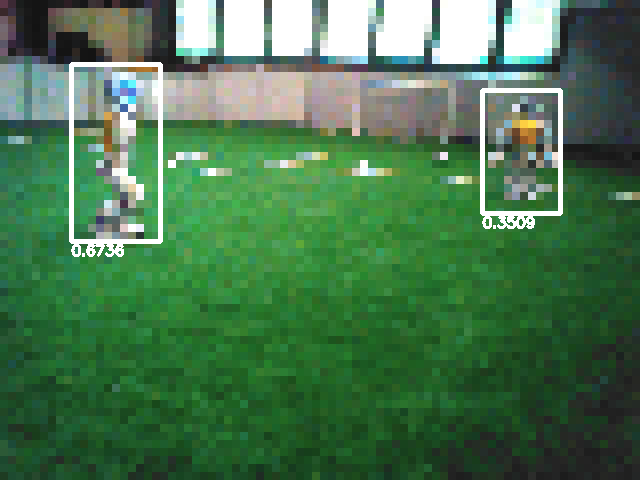}
	\caption{Two robots (Nao by SoftBank Robotics) detected in a 80x60 pixel image of  a soccer field.}
	\label{fig:robot_ex}
\end{figure}

We present above our ball detection application example, which is similar to the known R-CNN approach. Instead, for a robot detector application we build a pipeline based on the YOLO V2 approach~\cite{DBLP:conf/cvpr/RedmonF17} which is our third application example. In this paper we limit our presentation to the CNN utilized in the pipeline as described in Tab.~\ref{tab:robot}.

\subsection{Generic Deployment}

In this section we present different application scenarios and utilize NNCG, TensorFlow XLA and Glow to deploy executables including required steps to compile and link for the target platform. We study if the utilized tool is applicable under the circumstances of the scenario and show the simplicity by collecting the steps required for deployment.

\subsubsection*{Native Compilation for Host Platform}

This is the most simple scenario in this evaluation as all tools are able to generate code and compile natively. Additionally, source code and libraries for compilation of the tools are also available. The host is an Ubuntu 18.04.1 LTS 64 Bit.

NNCG generates a C source code file that can be compiled to an object file. There are no dependencies for the robot detection CNN except for SSSE3 intrinsics on Intel platforms (\verb|emmintrin.h|).
The ball and robot classification additionally depends on \verb|math.h| and \verb|libmath| caused by exponentional functions in Softmax. Thus, all ANSI C  compiler should be able to compile the C source file to an object file for a general architecture. Alternatively, if can be included in a project environment (CMake, Visual Studio etc.).

TensorFlow XLA includes the tool \verb|tfcompile| to generate object files from a trained and stored CNN. It thus includes one more step than NNCG, the compilation of the code utilizing clang.
However, the object file depends on many functions and the Eigen library shipped with TensorFlow. Thus, it is advisable to link this file to an executable within the TensorFlow environment providing all dependencies.

Glow's tool \verb|image-classifier| generates an object file utilizing clang as well. It does not depend on libraries as TensorFlow XLA making the linking process as easy as NNCG on this platform. However, as the output is an object, compilation is limited to platforms supporting clang. Furthermore, Glow does not support all layer required for a CNN based on the YOLO approach, namely leaky ReLU.

\subsubsection*{Deployment on Atom (J1900) with similar OS}

In this scenario the host platform for compilation is the same as above with a different target platform.
Two limitations differ this scenario. First, the target CPU only supports a limited subset of SIMD instruction compared to the host (SSSE3). Second, Ubuntu is installed in Version 14.04.5 LTS.

The C code file generated by NNCG can be compiled natively on the target platform as it only requires a basic C compiler installation. Alternatively, it can be compiled on the host with static linkage and by specifying the target architecture (\verb|bonnell| here).

TensorFlow XLA also supports the specification of a target platform. Static linkage is possible including the dependencies to TensorFlow and Eigen. However, a native compilation would require to install TensorFlow on the target platform and is thus not considered here.

Glow's \verb|image-classifier| does not allow to specify a different target platform. Thus, the generated object file contains AVX commands as these are available on the host but not on the target platform resulting in not working executables. Installation of Glow on the target platform was not considered here.

\subsubsection*{Deployment on Atom (Z530) with different OS}

This is the platform of the Nao robot with a preinstalled OS.
The CPU is more limited but supports the same SIMD extensions as above. Main difference here is the custom Linux distribution with 32 bit kernel. It does not provide a compiler, thus native compilation is impossible.

C source generated by NNCG can be cross compiled by specifying a 32 bit target and static linkage.
In contrast, the object generated by TensorFlow XLA depends on Eigen source that cannot be compiled for 32 bit targets. Glow suffers the same limitations as above and is not applicable here.

\subsection{Fast Executables}

The previous section demonstrates the applicability for different platforms. In this section we continue the evaluation by measuring the required time for the inference. We measure the time required to classify a single image (ball or pedestrian) and for detecting robots in an image. We ran small networks 100.000 times and larger networks like the robot detection 1000 times and use the mean value.
For each application example the results can be seen in Tab.~\ref{tab:ball_time}, ~\ref{tab:ped_time} and ~\ref{tab:nao_time}. As described in the previous section, some approaches are not applicable and thus no time measurement is available.

As can be seen, the speed-up factor of NNCG compared to TensorFlow XLA is between 1.41 and 11.81, compared to Glow 3.29. This also confirms the results of~\cite{rotem2018glow}. Additionally, we evaluated the ball and pedestrian CNN on a GeForce GTX 1050 GPU by NVIDIA using an executable by TensorFlow XLA. As can be seen, the overhead to utilize a GPU is tremendous for small CNN and does not change significantly for under 100 images classified at once.

As described in Sec.~\ref{sec:desprin}, two features of NNCG are configurable: the architecture dependent SIMD extensions and loop unrolling. We can therefore evaluate the acceleration due to these features by first using a general architecture without SIMD extensions where both outer loops are not unrolled. We do this for the ball classifier on the i7 platform. The compiler (clang 6.0.0) is nevertheless enabled to use SIMD extensions and perform loop unrolling. However, it can be seen in Tab.~\ref{tab:features} that the speed-up factor of applying SIMD instructions as described in Sec.~\ref{sec:simd} is 4.9. If NNCG unrolls all loops there is an additional speed-up of 26\%. This shows that the compiler is not able to find the optimum automatically.
\begin{table}[h]
	\caption{Execution time of ball classifier.}
	\label{tab:ball_time}
	\begin{center}
		\begin{tabular}{|c||c|c|c|}
			\hline
			Platform & NNCG & Glow & TensorFlow XLA\\
			\hline
			\hline
			Intel i7 (8650U) & 2.10\textmu s & 7.53\textmu s& 24.81\textmu s\\
			Intel Atom (J1900) & 17.51\textmu s & N/A & 69.12\textmu s\\
			Intel Atom (Z530) & 46.50\textmu s & N/A & N/A\\
			NVIDIA 1050 & N/A & N/A & 5630\textmu s\\
			\hline
		\end{tabular}
	\end{center}
\end{table}
\begin{table}[ht]
	\caption{Execution time of pedestrian classifier.}
	\label{tab:ped_time}
	\begin{center}
		\begin{tabular}{|c||c|c|c|}
			\hline
			Platform & NNCG & Glow & TensorFlow XLA\\
			\hline
			\hline
			Intel i7 (8650U) & 135.7\textmu s & N/A & 191.8\textmu s\\
			Intel Atom (J1900) & 1020.3\textmu s & N/A & 1757.2\textmu s\\
			Intel Atom (Z530) & 2938.6\textmu s & N/A & N/A\\
			NVIDIA 1050 & N/A & N/A & 5762\textmu s\\
			\hline
		\end{tabular}
	\end{center}
\end{table}
\begin{table}[H]
	\caption{Execution time of Robot detector.}
	\label{tab:nao_time}
	\begin{center}
		\begin{tabular}{|c||c|c|c|}
			\hline
			Platform & NNCG & TensorFlow XLA\\
			\hline
			\hline
			Intel i7 (8650U) & 474\textmu s & 2457\textmu s\\
			Intel Atom (J1900) & 1109\textmu s & 6797\textmu s\\
			\hline
		\end{tabular}
	\end{center}
\end{table}
\begin{table}[H]
	\caption{Speed comparison of different features.}
	\label{tab:features}
	\begin{center}
		\begin{tabular}{|c|c|c|}
			\hline
			General & SSSE3  & SSSE3 and Full Unroll\\
			12.94\textmu s & 2.64\textmu s & 2.10\textmu s\\
			\hline
		\end{tabular}
	\end{center}
\end{table}

\section{Conclusion and Outlook}
\label{sec:conclusion}
This paper presents a neural network code generator NNCG that writes ANSI C code for a trained CNN. We shown that embedding this file or a compiled object is a simple task and allows to deploy the CNN on all platforms that provide an ANSI C compiler or that can be a target platform of a cross compiler. Additionally, NNCG can exploit that the structure and gains of the CNN are know at generation time resulting in executables up to 11.81 times faster than previous well-known approaches.

Future work will focus GPU kernel code and more layer types to support modern widely known CNN structures. Furthermore, currently only SSSE3 is a supported SIMD instruction set. An extension of NNCG to other instruction sets like AVX or NEON can be realized rapidly.

\bibliographystyle{spmpsci}

\end{document}